# KAConvText: Novel Approach to Burmese Sentence Classification using Kolmogorov-Arnold Convolution

Ye Kyaw Thu[1,3]*   Thura Aung[2,3]*   Thazin Myint Oo[3]   Thepchai Supnithi[1]

[1]National Electronics and Computer Technology Center, Thailand
[2]King Mongkut's Institute of Technology Ladkrabang, Thailand
[3]Language Understanding Laboratory, Myanmar
{yekyaw.thu,thepchai.supnithi}@nectec.or.th
66011606@kmitl.ac.th   queenofthazin@gmail.com

## Abstract

This paper presents the first application of Kolmogorov-Arnold Convolution for Text (KAConvText) in sentence classification, addressing three tasks: imbalanced binary hate speech detection, balanced multiclass news classification, and imbalanced multiclass ethnic language identification. We investigate various embedding configurations, comparing random to fastText embeddings in both static and fine-tuned settings, with embedding dimensions of 100 and 300 using CBOW and Skip-gram models. Baselines include standard CNNs and CNNs augmented with a Kolmogorov-Arnold Network (CNN-KAN). In addition, we investigated KAConvText with different classification heads - MLP and KAN, where using KAN head supports enhanced interpretability. Results show that KAConvText-MLP with fine-tuned fastText embeddings achieves the best performance of 91.23% accuracy (F1-score = 0.9109) for hate speech detection, 92.66% accuracy (F1-score = 0.9267) for news classification, and 99.82% accuracy (F1-score = 0.9982) for language identification.

## 1 Introduction

Text classification is a key part of natural language processing (NLP). For low-resource languages like Burmese, this task is challenging due to issues such as unbalanced datasets, different dialects, and a lack of pre-trained language models. Traditional methods, like text classification with Convolutional Neural Networks (CNNs) (Kim, 2014), use linear transformations with fixed activation functions. While useful, these methods often have trouble capturing complex patterns in text, especially in languages with rich grammar and structure. This paper introduces a new approach to text classification using Kolmogorov-Arnold Convolution for Text (KAConvText) that uses spline-based non-linearities to improve flexibility and efficiency. We explore two inference approaches for the KAConvText models: one using a traditional Multi Layer Perceptron (MLP)-based softmax layer (KAConvText-MLP), and Kolmogorov-Arnold Network (KAN)-based approach (KAConvText-KAN).

Recent advancements in deep learning have seen the integration of advanced mathematical frameworks into neural architectures. Among these, KANs (Liu et al., 2024) stand out by leveraging the Kolmogorov-Arnold Representation Theorem (KART) (Schmidt-Hieber, 2021), (Selitskiy, 2022) to replace linear weight matrices with learnable spline functions. Bodner et al. (2024)[1] originally proposed the integration of the KAN concept with convolution for computer vision tasks.

In this paper, we adapted KAN principles to convolutional layers for text data, experimenting with KAConvText that replaces traditional one-dimensional CNN kernels with spline-parameterized functions. This innovation allows the model to dynamically learn non-linear mappings directly from data, circumventing the rigidity of fixed activation functions. Inspired by SplineCNN (Fey and Lenssen, 2018), which demonstrated splines' efficacy in geometric deep learning, our work extends these ideas with KAConvText to sequential text data, avoiding the need for graph-based preprocessing.

In this study, we trained and evaluated the proposed KAConvText models, with two different heads - MLP and KAN, across three Burmese sentence classification tasks: hate speech detection (imbalanced binary), news classification (balanced multiclass), and ethnic language identification (imbalanced multiclass). Our experiments demonstrate that the proposed KAConvText-MLP, when paired with fine-tuned fastText embeddings, achieve state-of-the-art performance: 91.23% accuracy (weighted F1-score: 0.9109) for hate speech detection, 92.66% accuracy (F1-score: 0.9267) for news categorization, and 99.82% accuracy (F1-score: 0.9982) for ethnic language identification. KAConvText-KAN, with enhanced interpretability, also achieves nearly identical best scores, with only minor score differences.

---

*Equal contribution.

[1]https://github.com/AntonioTepsich/Convolutional-KANs

These results highlight KAConvText's ability to perform well in both imbalanced and balanced classification while remaining interpretable in feature extraction and classification (with KAN classification layer).

## 2 Corpus Preparation

We prepared sentence classification datasets for three tasks: hate speech detection, news classification, and ethnic language identification. Table 1 shows the distributions of labels in eachs dataset. Each dataset was randomly split into an 80:20 ratio per class for training and testing.

For the news classification and hate speech detection experiments, we trained fastText embeddings (Bojanowski et al., 2017) using a cleaned, word-segmented Burmese monolingual corpus consisting of 341,221 sentences and 7,205,337 tokens. For the ethnic language identification task, we trained fastText embeddings on a syllable-segmented ethnic multilingual corpus containing 200,781 sentences and 2,755,734 tokens.

**Hate Speech Detection:** This imbalanced dataset comprises 10,140 annotated syllable-segmented sentences (8,493 hateful and 1,647 non-hateful), sourced from Myanmar social media platforms and public forums. Hate speech can have different categories such as abuse, religious bias, racism, body-shaming, political animosity, sexism, potentially lethal content, and educational bias (Kyaw et al., 2024b). For binary classification, we considered all categories of hate speech into a single "hate speech" class and retained the non-hateful content as the "non-hate speech" class.

**News Classification:** The news corpus contains 7,315 word-segmented news sentences evenly distributed across six categories. For news classification, we used the corpus, which includes six news categories - Sports, Politics, Technology, Business, Entertainment, and Environmental (Kyaw et al., 2024a).

**Ethnic Language Identification:** This dataset spans syllable-segmented nine ethnic language sentences which mostly share Burmese alphabets, totaling 108,016 sentences. Data was collected from the web sources and the previous machine translation researches [Oo et al. (2018), Oo et al. (2020a), Htun et al. (2021), Kyaw et al. (2020), Aye et al. (2020), Linn et al. (2020), Thu et al. (2019), Oo et al. (2020b), Oo et al. (2019), Myint Oo et al. (2019), Oo et al. (2020c), Thu et al. (2022)].

| Dataset | Class | Count | Percentage |
| --- | --- | --- | --- |
| Hate Speech | Hate | 8,493 | 83.76% |
| | Non-Hate | 1,647 | 16.24% |
| News | Sports | 1,232 | 16.84% |
| | Politics | 1,228 | 16.79% |
| | Technology | 1,224 | 16.73% |
| | Business | 1,221 | 16.69% |
| | Entertainment | 1,205 | 16.47% |
| | Environment | 1,205 | 16.47% |
| Language | Burmese | 19,519 | 18.07% |
| | Beik | 3,385 | 3.13% |
| | Dawei | 3,537 | 3.27% |
| | Mon | 5,854 | 5.42% |
| | Pa'o | 10,346 | 9.58% |
| | Po Kayin | 10,031 | 9.29% |
| | Rakhine | 9,778 | 9.05% |
| | S'gaw Kayin | 36,300 | 33.61% |
| | Shan | 9,266 | 8.58% |

Table 1: Label Counts and Percentages of Sentence Classification Datasets

## 3 Methodologies

Standard CNNs were used as baselines to isolate the impact of the proposed KAConv-Text architecture. We experimented with various embedding strategies—random initialization and pretrained fastText embeddings in both static and fine-tuned modes. Evaluated models include standard CNNs, CNN-KAN variants, and the proposed KAConvText. Model architectures are shown in Figure 1.

### 3.1 Embeddings

Word embeddings map discrete tokens to continuous vectors, capturing semantic and syntactic relationships. We evaluate two initialization strategies - random embeddings and pre-trained embeddings.

In random embeddings, tokens are assigned random vectors, which are learned during training. This approach requires no prior linguistic knowledge but demands sufficient training data to learn meaningful representations. The second approach

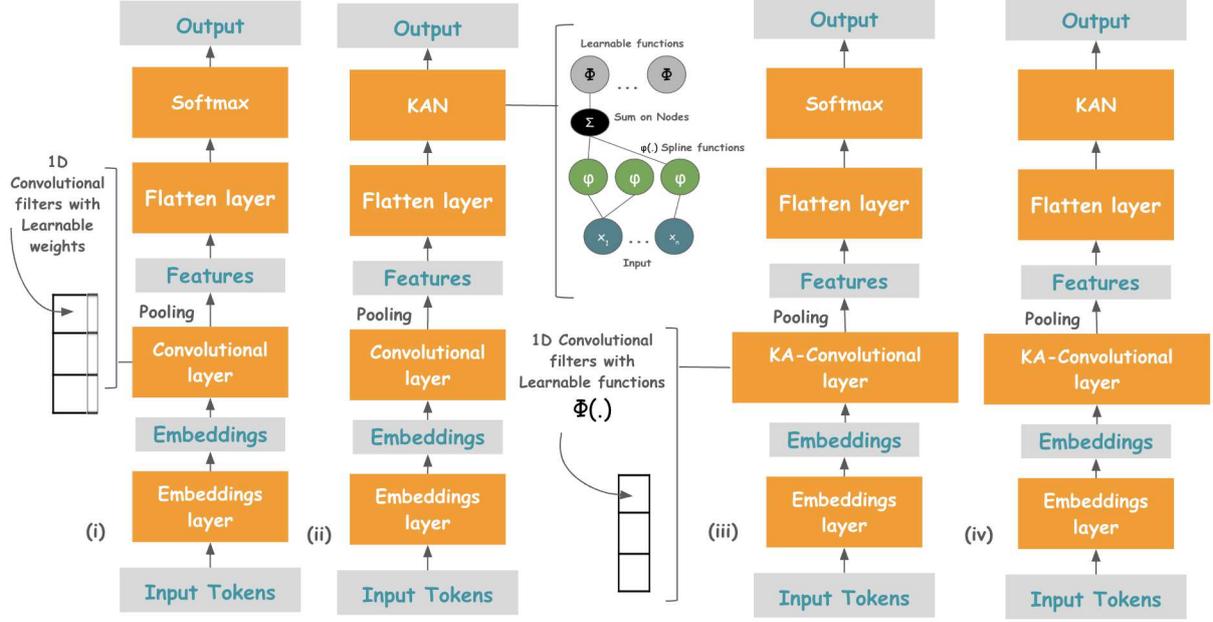

Figure 1: Architectural variations of experimented models: (i) Standard CNN, (ii) CNN-KAN with classification layer adaptation, (iii) KAConvText-MLP and (iv) KAConvText-KAN

employs fastText embeddings, pre-trained on our monolingual corpus (see Section 2), with two settings: static and fine-tuned. In the static setting, fastText vectors remain fixed during training. In the fine-tuned setting, the vectors are updated through backpropagation, allowing them to adapt to task-specific contexts.

### 3.2 Kolmogorov-Arnold Networks

#### 3.2.1 Kolmogorov-Arnold Representation Theorem (KART)

KART (Kolmogorov, 1956) simplifies complex multivariate functions by decomposing them into simpler univariate functions. Given a function $f(x_1, x_2, \ldots, x_n)$ that produces a single output, the theorem shows that any smooth function can be represented by first processing each input $x_p$ with simple single-variable functions $\phi_{q,p}$. The outputs of these functions are then summed, and a final adjustment is made by applying another set of single-variable functions $\Phi_q$ to these sums.

This framework highlights that complex interactions in multivariate functions can be effectively captured through a composition of simple, univariate transformations and additive operations, significantly reducing the complexity of representing high-dimensional mappings. Mathematically, this is written as:

$$f(\mathbf{x}) = \sum_{q=1}^{2n+1} \Phi_q \left( \sum_{p=1}^{n} \phi_{q,p}(x_p) \right) \quad (1)$$

where $\phi_{q,p}$ are simple functions that handle each input $x_p$ individually and $\Phi_q$ are functions that adjust the combined results to reconstruct $f(\mathbf{x})$. Instead of using fixed activation functions (like ReLU), KANs use these adaptive $\phi_{q,p}$ and $\Phi_q$ functions to learn complex patterns.

#### 3.2.2 KAN Architecture

Kolmogorov-Arnold Networks (KANs) extend the classical Kolmogorov-Arnold representation by using learnable activation functions along graph edges. In a KAN layer, an input with $n_{\text{in}}$ dimensions is transformed into an output with $n_{\text{out}}$ dimensions. This layer is defined by a collection of functions: $\Phi = \{\phi_{q,p}\}$, $p = 1, 2, \ldots, n_{\text{in}}$, $q = 1, 2, \ldots, n_{\text{out}}$, where each $\phi_{q,p}$ is a learnable spline function.

The output of the layer is computed by applying these functions to the input features. For the $l$th layer, the output is given by:

$$\mathbf{x}^{(l+1)} = \sum_{i=1}^{n_l} \phi_{l,j,i}(x_i^{(l)}), \quad j = 1, \ldots, n_{l+1}. \quad (2)$$

In matrix notation, this operation is written as:

$\mathbf{x}^{(l+1)} = \Phi^{(l)} \mathbf{x}^{(l)}$, where $\Phi^{(l)}$ is the function matrix for the $l$th layer.

In particular, the approximation error for KANs is bounded by

$$\left\| f - \left( \Phi_G^{(L-1)} \circ \cdots \circ \Phi_G^{(0)} \right) \mathbf{x} \right\|_{C^m} \leq C\, G^{-k-1+m}, \quad (3)$$

where $G$ is the grid size, $k$ is the order of the B-spline, and $C$ is a constant.

The notation $\Phi_G^{(L-1)} \circ \cdots \circ \Phi_G^{(0)}$ indicates that the output of one function matrix is fed as the input to the next.

### 3.3 Integrating with Convolution

#### 3.3.1 Convolutional Neural Networks

Convolutional Neural Networks (CNNs) were originally built for visual tasks with two-dimensional convolutions (LeCun et al., 1989). In our paper, the baseline CNN architecture processes embeddings $\mathbf{X}$ with one-dimensional convolutions, followed by a pooling layer and a linear classifier. For a kernel $\mathbf{W} \in \mathbb{R}^{k \times d \times C}$ with $C$ output channels:

$$\mathbf{H}_c = \mathrm{ReLU}\left( \sum_{m=1}^{k} \sum_{n=1}^{d} \mathbf{W}_{c,m,n} \cdot \mathbf{X}_{i+m-1,n} + b_c \right) \quad (4)$$

where $b_c$ is the bias term. The final feature representation is pooled and classified via softmax function: $\hat{y} = \mathrm{Softmax}(\mathbf{W}_{\mathrm{cls}} \mathbf{h} + \mathbf{b}_{\mathrm{cls}})$.

#### 3.3.2 KAN as classification layer

CNN-KAN approach was introduced shortly after the development of Kolmogorov-Arnold Networks (KAN) to enhance their performance in visual tasks (Cang et al., 2024). CNN-KAN models combine convolutional feature extraction with the expressive, spline-based nonlinearity of Kolmogorov-Arnold Networks (KAN) to better handle spatially structured visual data. In this paper, standard one-dimensional convolutional layers first perform feature extraction, producing feature maps that capture local text patterns. These feature maps are then passed through a KAN-inspired activation, where the conventional linear transformation is augmented with spline-parameterized non-linearities.

#### 3.3.3 Kolmogorov-Arnold Convolution

The idea of KAN is extended to one-dimensional convolution by replacing the standard dot product with a learnable nonlinear function defined via B-splines. In this approach, unlike its use in Computer Vision (Azam and Akhtar, 2024), a KAConvText's spline kernel is represented as Kernel $= [\phi_1, \phi_2, \ldots, \phi_K]$ where each $\phi_m$ is a univariate non-linear function. Following the original proposal (Liu et al., 2024), each learnable function is defined by

$$\phi(x) = w_s \cdot \mathrm{spline}(x) + w_b \cdot b(x), \quad (5)$$

where $b(x)$ is a fixed basis function like Parametric ReLU (PReLU) and $\mathrm{spline}(x)$ is a linear combination of B-spline basis functions. The learnable parameters $w_s$ and $w_b$ are a scalar gate to $b(x)$ and $\mathrm{spline}(x)$ respectively. They control how much of the basis versus the spline part contributes. The extracted feature representation is pooled, flattened, and classified using MLP-based softmax layers for **KAConvText-MLP**. For the **KAConvText-KAN**, KAN layer is used as a classification layer.

#### 3.3.4 Visualizability

**CNN-KAN** improves interpretability by replacing the final linear classifier with a KAN layer. The spline-based activation functions in KAN are learnable and can be visualized as univariate functions, revealing how input features are nonlinearly transformed before classification. Unlike fixed activations in CNNs, these splines adapt to the data, allowing to be inspected how specific feature ranges influence class probabilities. The convolutional layers in CNN-KAN remain unchanged, retaining CNN's lack of transparency in feature extraction. As a result, while the classification stage becomes more interpretable, the extraction of intermediate positional patterns remains unclear. KAConvText models - **KAConvText-MLP** and **KAConvText-KAN** enhance interpretability at the convolutional level by replacing linear kernels with learnable spline functions. Each kernel operates as a transparent, univariate nonlinear transformation whose B-spline parameters can be directly visualized to understand its response to input. KAConvText-KAN, in particular, offers enhanced interpretability by ensuring that both feature extraction and classification rely on transparent, learnable spline-based transformations.

Figure 2 visualizes the learned B-spline surfaces for each KAConvText kernel, where the X-axis represents spline coefficients, the Y-axis denotes individual filters, and the Z-axis reflects weight values, thereby offering an intuitive view of how these nonlinear transformations shape feature responses. Early in training (Epoch 1) we can find tall, sharp spikes—this is because the model is still exploring and some spline bases briefly jump to large values. By Epoch 10 the overall height of the surface is higher (magnitude ↑) but the spikes have smoothed out (spikiness ↓). The model has found which spline patterns are useful and turned those up so they matter more. Spline curves and weight decay work together to flatten the extreme peaks, gradually smoothing the surface as the model converges on its final feature representations.

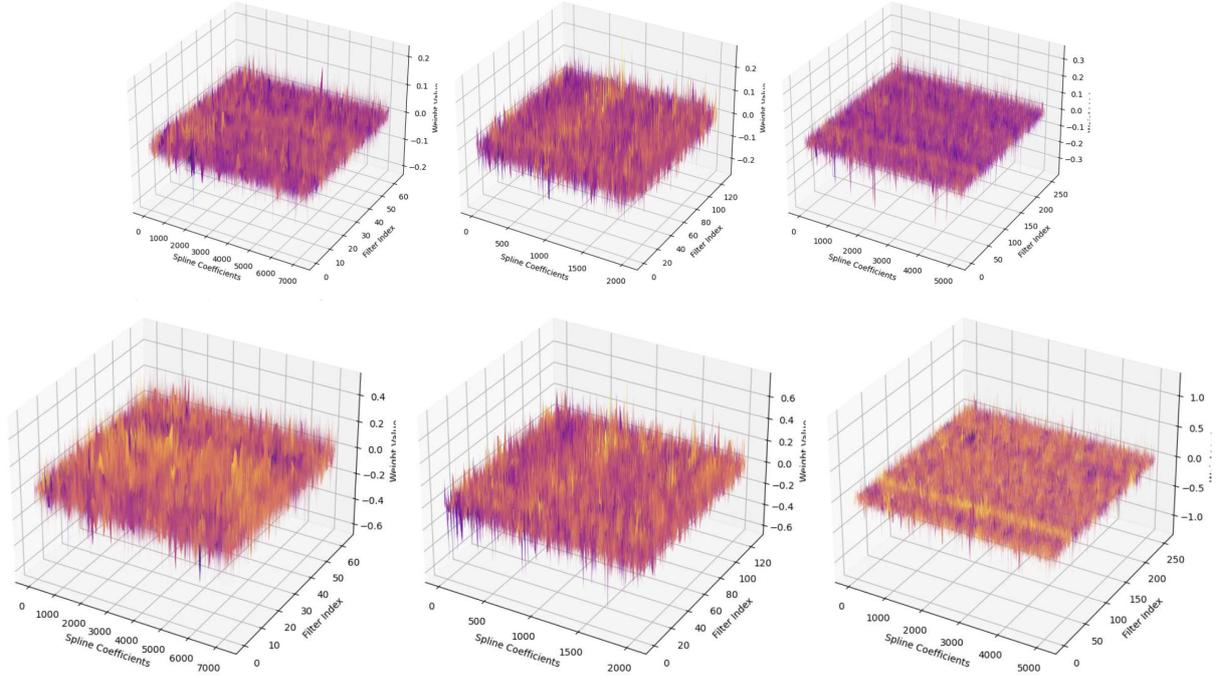

Figure 2: 3D visualization of the learned B-spline surfaces for each KAConvText convolutional kernel in the fine-tuned fastText (300 dimensions, Skip-gram) + KAConvText-MLP model for News Classification. Top: Epoch 1. Bottom: Epoch 10. The color scale ranges from smaller to larger weight values, transitioning from bright purple to yellow, starting at mid-brightness.

## 4 Experimental Setup

In this study, we explore a range of embedding strategies, including randomly initialized vectors and pre-trained fastText embeddings. These embeddings are evaluated in both static and fine-tuned configurations, with dimensionalities of 100 and 300. The fastText embeddings are trained using the CBOW and Skip-gram algorithms, and the best performing results are compared. For training, 80% of each dataset is used while 20% of data is used for validation and testing.

We trained and evaluated four models: CNN, CNN-KAN, KAConvText-MLP and KAConvText-KAN. The models were evaluated using accuracy and weighted F1-scores. For each model, we employ a three-layer architecture with successive channel dimensions of 64, 128, and 256. The kernel sizes for these layers are set to 3, 4, and 5, respectively. ReLU activation is applied after each convolutional layer for CNN and CNN-KAN models, and an adaptive average pooling operation is used for pooling. In the case of KAConvText models, we incorporate cubic splines for the convolutions, with the polynomial degree of the B-spline basis functions set to three. All models were trained for 10 epochs with a dropout rate of 0.3. Instance Normalization and GELU[2] are used before base convolution in KAConv layers. Table 2 presents other hyperparameters used in the KAN and KAConvText models.

For all models, the Adam optimizer was used for model optimization, with a learning rate of 1e-3. For model training, we utilized the Kaggle environment[3], leveraging an NVIDIA P100 GPU with 16GB of memory for efficient computation. The experiments were conducted using PyTorch[4] framework developed by Paszke et al. (2019).

---
[2] https://pytorch.org/docs/stable/generated/torch.nn.GELU.html
[3] https://www.kaggle.com/
[4] https://pytorch.org/

| Hyperparameter | Description |
| --- | --- |
| grid size | number of spline intervals (default: 5) |
| spline order | B-spline order (default: 3) |
| scale noise + | noise scale for initializing spline weights (default: 0.1) |
| scale base + | scaling factor for base (linear) weights (default: 1.0) |
| scale spline + | scaling factor for spline weights (default: 1.0) |
| grid eps + | blend between adaptive and uniform grid (default: 0.02) |
| grid range | value range of spline grid (default: $\{-1, 1\}$) |
| stride * | convolution stride (default: 1) |
| padding * | convolution padding (default: 0) |
| dilation * | convolution dilation (default: 1) |
| groups * | number of groups in conv (default: 1) |

Table 2: Hyperparameters: "+" only used for KAN, "*" only used for KAConvText models, and others for both. Default values are shown in parentheses.

## 5 Results and Discussion

This section analyzes how different types of embeddings—random, static, and fine-tuned—affect the performance of each model architecture across various tasks. We compare the models based on their accuracy and weighted F1-scores, highlighting the best-performing configurations for each embedding type. Additionally, we discuss the efficiency and computational complexity of the models, providing insights into the trade-offs between performance and resource usage.

### 5.1 Embedding Impact on Model Performance

**Hate Speech Detection** Figure 3 (a) indicate that fine-tuned embeddings again lead to improved performance. In CNN, moving from random to static embeddings yields a large jump (F1-score: 0.8895 → 0.9001). Fine-tuning adds a smaller gain (to 0.9023), suggesting most of CNN's improvement comes from pre-training rather than task-specific adaptation. For CNN-KAN, the static Skip-gram embeddings boost performance markedly over random (F1-score: 0.8851 → 0.9017). Surprisingly, fine-tuning CBOW slightly degrades performance to 0.8946, indicating that the flexible spline activations in CNN-KAN may already adapt sufficiently to the task. For KAConvText models, while training with MLP inference, at 100 dimension, static embeddings give modest gains (+0.0036 over random for CBOW), and fine-tuning further adds only +0.0035. At 300 dimension, static Skip-gram yields a +0.0230 jump, while fine-tuned Skip-gram delivers the best F1-score of 0.8931—highlighting that larger, tuned embeddings best exploit the spline-based feature extractor. KAConvText-KAN's performance follows the same pattern as KAConvText-MLP: static Skip-gram at 300 dimension improves over random, and fine-tuned Skip-gram edges it further to F1 0.8931. However, it remains slightly below KAConvText-MLP at each setting, suggesting the MLP head can better leverage the learned representations.

**News Classification** Figure 3 (b) reveal that fine-tuned embeddings consistently yield superior results. In CNN model, the static embeddings alone raise F1-score 0.8823 to 0.9124; fine-tuning Skip-gram at 100 dimension pushes it to 0.9196. This underscores that pre-training provides the bulk of the benefit, with fine-tuning yielding diminishing returns. For CNN with KAN head, similar to CNN, static Skip-gram gives +0.0202 (0.8932 → 0.9134) and fine-tuning CBOW improves another +0.0045. CNN-KAN's spline activations do not fundamentally alter the embedding effect. For KAConvText models, while training with MLP inference, at 100 dimension, static Skip-gram underperforms CBOW, but fine-tuned CBOW recovers to 0.9022. At 300 dimension, static Skip-gram delivers a massive +0.1155 gain, and fine-tuned Skip-gram tops out at F1-score of 0.9185, confirming that high-dimensional pre-trained embeddings are most effective. For KAConvText-KAN, The best result (0.9185) comes from 300 dimension fine-tuned Skip-gram, with static Skip-gram (0.9122) close behind—again indicating the classifier head has limited influence on embedding-driven gains.

**Language Identification** Figure 3 (c) demonstrate near-perfect performance across all models. The highest scores were observed with the 300-

| Embed | Model | Hate Speech | | News | | Language | |
|---|---|---|---|---|---|---|---|
| | | Acc (%) | F1 | Acc (%) | F1 | Acc (%) | F1 |
| Rand | CNN | **88.96** | **0.8895** | 88.17 | 0.8823 | **99.73** | **0.9973** |
| | CNN-KAN | 88.81 | 0.8851 | 89.26 | 0.8932 | **99.73** | **0.9973** |
| | KAConvText-MLP | 88.62 | 0.8819 | **89.67** | **0.8969** | 99.71 | 0.9971 |
| | KAConvText-KAN | 87.19 | 0.8719 | 79.61 | 0.7967 | 99.62 | 0.9962 |
| Static | CNN | 89.90 | 0.9001 | 91.23 | 0.9124 | 99.71 | 0.9971 |
| | CNN-KAN | **90.29** | **0.9017** | 91.30 | 0.9134 | 99.72 | 0.9972 |
| | KAConvText-MLP | 89.16 | 0.8855 | **91.50** | **0.9147** | 99.73 | 0.9973 |
| | KAConvText-KAN | 89.01 | 0.8809 | 91.16 | 0.9122 | **99.75** | **0.9975** |
| Fine-tuned | CNN | 90.24 | 0.9023 | 91.98 | 0.9196 | 99.78 | 0.9978 |
| | CNN-KAN | 89.40 | 0.8946 | 91.77 | 0.9179 | 99.79 | 0.9979 |
| | KAConvText-MLP | **91.23** | **0.9109** | **92.66** | **0.9267** | **99.82** | **0.9982** |
| | KAConvText-KAN | 89.16 | 0.8931 | 91.77 | 0.9185 | 99.76 | 0.9976 |

Table 3: Accuracy (Acc) and Weighted F1-scores (F1-score of the best models across different embeddings settings on Hate Speech Detection (Binary), News (Multiclass), and Language Identification (Multiclass) classification tasks. Bolded results are the best result for each task across each embedding type.

| Task | Model | Train Time (sec) | Eval Time (sec) | Parameter Count |
|---|---|---|---|---|
| Hate Speech | CNN | 26.19 | 0.23 | 958,070 ($\sim$ 1.0M) |
| | CNN-KAN | 45.20 | 0.39 | 962,676 ($\sim$ 1.0M) |
| | KAConvText-MLP | 188.74 | 2.63 | 2,991,289 ($\sim$ 3.0M) |
| | KAConvText-KAN | 207.68 | 2.69 | 2,995,895 ($\sim$ 3.0M) |
| News | CNN | 20.75 | 0.17 | 3,372,298 ($\sim$ 3.4M) |
| | CNN-KAN | 34.28 | 0.29 | 3,386,116 ($\sim$ 3.4M) |
| | KAConvText-MLP | 141.02 | 1.96 | 5,405,517 ($\sim$ 5.4M) |
| | KAConvText-KAN | 154.74 | 2.00 | 5,419,335 ($\sim$ 5.4M) |
| Language | CNN | 663.78 | 2.51 | 4,684,669 ($\sim$ 4.7M) |
| | CNN-KAN | 1052.81 | 4.25 | 4,705,396 ($\sim$ 4.7M) |
| | KAConvText-MLP | 4136.40 | 29.89 | 6,717,888 ($\sim$ 6.7M) |
| | KAConvText-KAN | 4418.59 | 30.83 | 6,738,615 ($\sim$ 6.7M) |

Table 4: Comparison of model train and evaluation time, and total trainable parameter count. Fine-tuned fastText embeddings with 300 dimensions for this comparison.

dimensional fine-tuned Skip-gram setting, where the CNN model reached an F1-score of 0.9978 and the KAConvText with MLP head achieved 0.9982.

## 5.2 Comparison of Best Models Across Each Setting

The performance of CNN, CNN-KAN, and KAConvText models across the three classification tasks—Hate Speech Detection, News Classification, and Language Identification—is presented in Table 3. We compare how each model responds to different embedding settings (random, static, and fine-tuned), considering both F1-score and accuracy.

**Random embeddings** For **Hate Speech Detection**, the CNN model achieved the highest F1-score (0.8895) and accuracy (88.96%), slightly outperforming other models. This suggests that CNN is sufficient for extracting discriminative features even with untrained embeddings. In **News Classification**, KAConvText-MLP stood out with the best F1-score (0.8969) and accuracy (89.67%), showing its ability to handle multiclass tasks better in the absence of pretrained embeddings. For **Language Identification**, all models performed exceptionally well (99.62% Acc), with CNN and CNN-KAN both achieving a high of 0.9973 F1-score.

**Static Embeddings** Compared to using random embeddings, performance improves across all tasks.

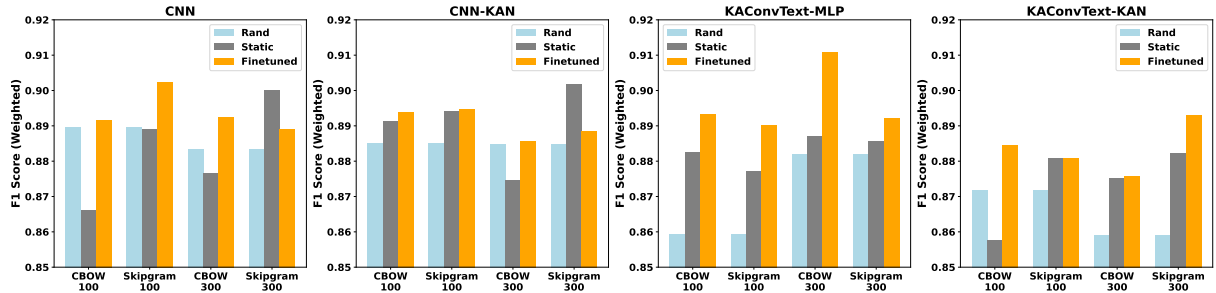

(a) F1-scores for Hate Speech Detection Task

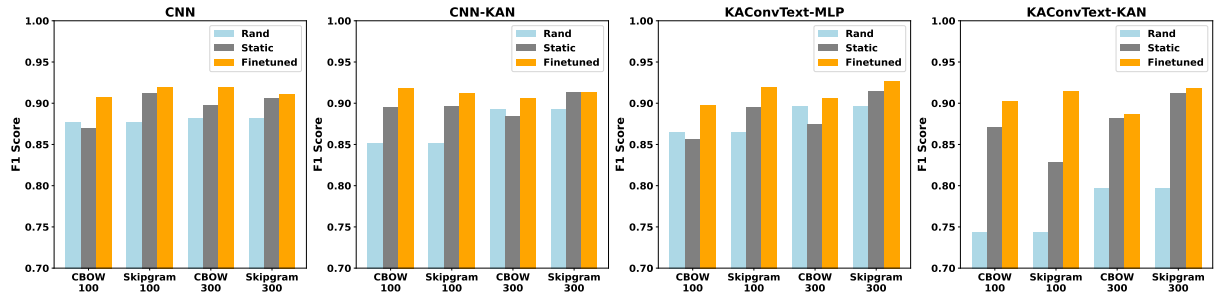

(b) F1-scores for News Classification Task

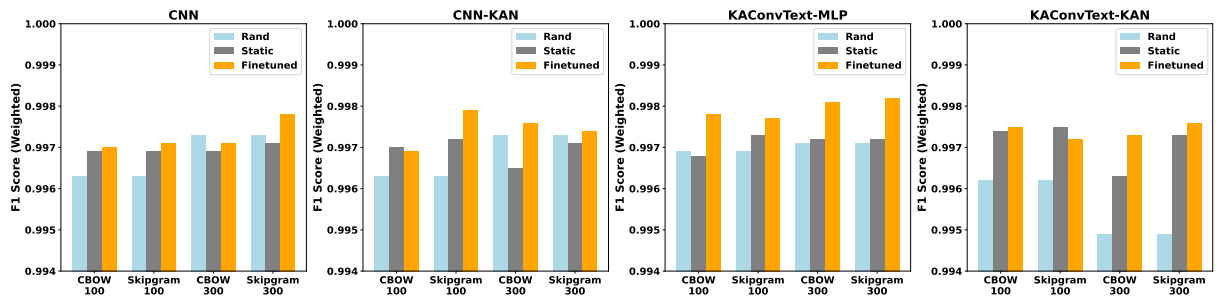

(c) F1-scores for Language Identification Task

Figure 3: F1 Scores of Different Embeddings Settings - Dimensions (100, 300) and Algorithms (CBOW, Skip-gram) for Each Task. The figure shows the importance of finetuning and selecting appropriate embedding dimensions for maximizing model performance across different tasks. We selected best performing settings for each model across each task for further evaluation.

In **Hate Speech Detection**, CNN-KAN slightly edges out other models with 90.29% accuracy and 0.9017 F1, indicating that KAN's learned attention over convolutional features adds value even when the embeddings are fixed. For **News Classification**, KAConvText-MLP achieves the best result (91.50% Acc, 0.9147 F1-score, suggesting that combining KAConvText and MLP classification layer with fixed embeddings is effective for capturing topical distinctions. **Language Identification** again sees near-perfect performance across models, with KAConvText-KAN scoring the highest (0.9975 F1-score, but overall differences remain minimal (<0.004), confirming the simplicity of the task.

**Fine-Tuned Embeddings** Using fine-tuning yield the **highest gains** across tasks. In **Hate Speech Detection**, KAConvText-MLP reaches the best performance with 91.23% accuracy and 0.9109 F1, improving over CNN by nearly +1% in accuracy. This highlights the benefit of fine-tuning when combined with a deeper, adaptive architecture. In **News Classification**, KAConvText-MLP again leads (92.66% Acc, 0.9267 F1-score, demonstrating superior capability in capturing semantic variation and subtle topic boundaries. For **Language Identification**, all models are near ceiling performance. KAConvText-MLP achieves the highest F1-score (0.9982), but the difference from other models is negligible (<0.1%), reflecting the limited room for improvement in this task.

## 5.3 Efficiency and Complexity

Table 4 shows the comparison of models with fine-tuned 300-dim fastText embeddings training time, evaluation time, and the total number of trainable parameters across different model architectures, including the embedding layer. As expected, models with CNN feature extraction exhibit the fastest training and inference times with the smallest parameter counts, making them the most efficient. Using KAConvText increases both training time and parameter count, reflecting the added complexity of spline-based transformations.

The training and evaluation time difference between using CNN and KAConvText is more significant when the data size is bigger. KAConvText variants consistently have higher parameter counts and longer runtimes, particularly in the Language Identification task, where the training time exceeds 4,000 seconds for KAConvText-KAN. This illustrates the trade-off between computational cost and representational power, with KAConvText-KAN being the most complex yet potentially expressive among the models evaluated.

## 6 Conclusion and Future Work

In this work, we present the first study to explore the use of KAConvText layers in text classification. Our main contributions include: (1) newly cleaned and curated Burmese text classification datasets, and (2) two novel KAConvText models—KAConvText-MLP and KAConvText-KAN—which outperform standard CNN-based architectures across multiple tasks. Our results show that combining KAConvText layers with fine-tuned fastText embeddings consistently improves classification performance.

In particular, the KAConvText-MLP model achieves the best results overall, while KAConvText-KAN, which is more interpretable, demonstrates competitive performance with a fully KAN-integrated structure. We also found that models with CNN feature extraction are the most efficient in terms of time and parameters, while KAConvText variants trade off efficiency for increased complexity and representational power.

In the future, we plan to extend our evaluation across a broader range of languages and datasets to assess cross-linguistic and cross-domain generalizability, including low-resource settings. We also intend to explore the integration of contextual embeddings (e.g., BERT or LLM-derived features) with KAConvText, and to investigate additional KAN-based architectures. Furthermore, we aim to expand our work beyond single-instance classification by applying KAConvText to pairwise classification tasks, enabling applications such as textual similarity, paraphrase detection, and natural language inference.